\documentclass[11pt]{article}

\usepackage{booktabs} 
\usepackage{url}  
\usepackage{graphicx}  
\usepackage{booktabs} 
\usepackage{luca} 
\usepackage{graphicx}
\usepackage{amsfonts,amsmath,amssymb}
\usepackage{url}
\usepackage{hyperref}
\usepackage{comment} 
\usepackage{float} 
\usepackage{tablefootnote} 
\usepackage{xcolor}
\usepackage{geometry} 
\usepackage{lipsum}
\usepackage{algorithm}
\usepackage{algorithmic}
\usepackage{ourmath}

\begin{document}

\title{Learning Edge Properties in Graphs from Path Aggregations\footnote{To be published in The Proceedings of the 2019 World Wide Web Conference (WWW'19)}}

	\author{Rakshit Agrawal \quad Luca de Alfaro 
	\\[1ex]
	\normalsize Computer Science and Engineering Department\\
	\normalsize University of California, Santa Cruz\\
	\normalsize ragrawa1@ucsc.edu, luca@ucsc.edu}
\date{March 11, 2019 \\
}
\maketitle

\begin{abstract}
Graph edges, along with their labels, can represent information of fundamental importance, such as links between web pages, friendship between users, the rating given by users to other users or items, and much more. 
We introduce LEAP, a trainable, general framework for predicting the presence and properties of edges on the basis of the local structure, topology, and labels of the graph. 
The LEAP framework is based on the exploration and machine-learning aggregation of the paths connecting nodes in a graph. 
We provide several methods for performing the aggregation phase by training path aggregators, and we demonstrate the flexibility and generality of the framework by applying it to the prediction of links and user ratings in social networks. 

We validate the LEAP framework on two problems: link prediction, and user rating prediction.
On eight large datasets, among which the arXiv collaboration network, the Yeast protein-protein interaction, and the US airlines routes network, we show that the link prediction performance of LEAP is at least as good as the current state of the art methods, such as SEAL and WLNM. 
Next, we consider the problem of predicting user ratings on other users: this problem is known as the edge-weight prediction problem in weighted signed networks (WSN).
On Bitcoin networks, and Wikipedia RfA, we show that LEAP performs consistently better than the Fairness \& Goodness based regression models, varying the amount of training edges between 10 to 90\%.
These examples demonstrate that LEAP, in spite of its generality, can match or best the performance of approaches that have been especially crafted to solve very specific edge prediction problems. 
\end{abstract}

\section{Introduction}
\label{sec-introduction}

Graphs and networks provide the natural representation for many real-world systems and phenomena.
In social networks, for instance, different users can be represented by nodes in a graph, and each friendship relation can be represented as an edge.
Similarly, in physical systems such as railroad networks or communication networks, each terminal or cellular station is a node in the graph connected to several other terminals.
In these graphs, each edge can itself contain significant amount of information.
For instance, the presence of an edge in a social network gives a binary signal of presence or absence of a relationship, and a weighted edge on the same network can give a numerical measure of the relationship, hence increasing the degrees of available information.
A signed network further contains ``positive'' and ``negative'' edge weights, indicating the intensity as well as the direction of a relationship.

\begin{figure}[tb]
	\centering
	\includegraphics[width=1.0\columnwidth]{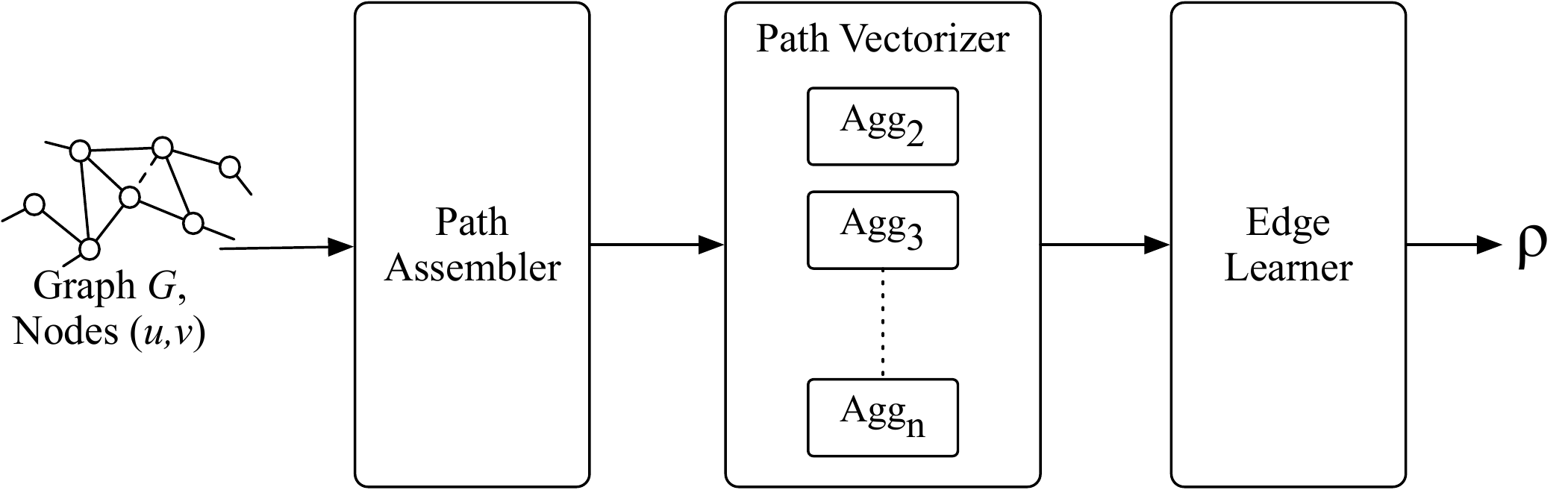}
	\caption{Summarized architecture of LEAP framework}
	\label{fig-summary}
\end{figure}

The presence and properties of edges in graphs are influenced by several structural factors such as the local neighborhood of the edge, the topology of the graph, and properties and labels associated with surrounding edges in the graph, among others.
Machine learning methods can be used to predict the existence of edges, or their properties. 
For instance, the problem of link prediction is a well explored research area where machine learning methods ranging from heuristics to deep neural networks have been experimented.
Similarly, problems around predicting specific edge properties or weights in a graph can also be addressed using learning algorithms.

In this paper, we present a general deep learning framework for learning and predicting edge properties in graphs on the basis of the local neighborhood of the edges.
The framework uses the concept of aggregating paths in the graph and is named LEAP (Learning Edges by Aggregation of Paths).
A distinctive feature of LEAP is its generality: its ability to learn any kind of edge properties without special need for feature extraction.

In LEAP, for a given graph $G=(V,E)$ where $V$ is the set of nodes and $E$ is the set of edges in the graph, and any two given nodes $(u,v) \in V$, we aim at predicting properties associated with the edge $e_{u,v}$ between the two nodes.
For instance, in link prediction, we aim at predicting the presence or absence of this edge, whereas in edge weight prediction, our objective is to predict the weight $w_{u,v}$ associated with the edge $e_{u,v}$.
LEAP has a modular architecture consisting of three primary modules: path assembler, path vectorizer, and edge learner (Figure~\ref{fig-summary}).
Combined together, these three modules form an end-to-end trainable system.
The path assembler gathers paths of different lengths between the node pair $(u,v)$.
The pact vectorizer uses an aggregator to summarize the information on the paths into a single vector representation.
The edge learner uses the vector representation derived from the path vectorizer, and learns a specific objective for any given edge property.

LEAP's modular architecture makes it easy to implement and experiment.
The path aggregators used in LEAP are deep learning modules that take as inputs the raw features of the paths, and produce a trainable aggregation, which is akin to an embedding of the edge set into a vector space.
The aggregators thus perform --- automatically by computing an embedding --- the feature engineering that used to be performed manually, for each property of interest (edge prediction, edge weight prediction, and so on). 
LEAP, while being an end-to-end learning system that learns embeddings for the nodes and paths itself, can also use pre-trained node embeddings \cite{Perozzi2014DeepWalk,grover2016node2vec}, node features, and edge features whenever available.

We present four different kinds of aggregators for LEAP.
The aggregators, AvgPool, DenseMax, SeqOfSeq, and EdgeConv, use different neural components and operate at different levels of complexity, focusing on properties like the ordered nature of nodes in a path, and the properties of edges in the paths.
We use standard neural modules such as Long Short-Term memory (LSTM) \cite{Hochreiter1997LongSM} networks, Convolutional Neural Networks (CNN) \cite{Bengio95convolutionalnetworks}, Pooling operations (Max and Average), and Feed-forward neural networks while constructing our aggregators.

We validate our framework on two specific graph problems: link prediction \cite{libenLinkPredictionProblem}, and edge weight prediction in weighted signed networks (WSN)~\cite{KumarWSN}.
In link prediction, we evaluate LEAP on eight real world datasets and present comparisons on the Area under the ROC curve (AUC) score with the current state of the art models WLNM~\cite{zhang2017wlnm} and SEAL \cite{zhang2018seal}, and more baseline methods.
In the WSN edge weight prediction task presented in Kumar et al.\ \cite{KumarWSN} , we evaluate LEAP on three user-user interaction datasets. 
Two of these datasets refer to Bitcoin trading networks where users provide a rating to other users based on trust.
In the third dataset, we learn weights for the votes and sentiment scores assigned by users in Wikipedia to other users when one submits a request for adminship (RfA).
We show that LEAP performs similar or better on both these problems against dedicated methods crafted for the specific problems.  

The primary contributions of this work can be summarized as follows:
\begin{itemize}
	
	\item
	We present and implement a novel deep learning framework, LEAP,  for learning and predicting edge properties of graphs. 
	The framework is general, and it requires no feature engineering, as it relies on deep learning to predict edge properties.
	
	\item
	We define several edge aggregators for LEAP, each suited to particular classes of prediction problems, and we illustrate how LEAP can take advantage of any graph embeddings that may be already available.

	\item We consider two standard graph prediction problems: link prediction (used, e.g., to predict the formation of connections in social networks) and edge weight prediction (used, e.g., to predict user ratings).  
	We show that LEAP, in spite of its generality, closely matches or improves, on the performance of specialized systems that have been built for these tasks.
	
\end{itemize}

In the paper, we will first discuss some related methods for edge property prediction in graphs.
We then discuss the motivation behind our LEAP framework and the usage of paths.
This is followed by the system design and detailed discussion on aggregators.
We then present results from an extensive evaluation over several datasets.
We conclude the paper with some considerations on the extensibility and modular design of LEAP.

\section{Related Work}
\label{sec-background}

Graphs have generated intense research interest over the years in machine learning problems.
Commonly studied problems include link prediction \cite{libenLinkPredictionProblem}, node classification, and node ranking \cite{Shahriari2014RankingNI}, among others.

With growing interest in deep learning for graphs, several algorithms for learning node representations have been suggested.
These include embedding methods such as LINE \cite{Tang2015LINELI}, DeepWalk \cite{Perozzi2014DeepWalk} and node2vec \cite{grover2016node2vec}.
More neural network based methods for learning node representations include Graph Convolutional Networks \cite{kipf2017semi}, GraphSAGE \cite{hamilton2017inductive}, and Graph Attention Networks \cite{gat2018}.
These methods are also often adapted for edge-based learning tasks such as link prediction.

Link prediction has been performed with methods ranging from heuristics to deep neural networks.
Martinez \emph{et al.} \cite{MartinezLPSurvey} have categorized the existing link prediction methods into the similarity based, probabilistic and statistical, algorithmic, and preprocessing categories.
Similarity based methods operate on the intuition of similar nodes having an affinity towards each other.
Locally focused methods like Common Neighbors \cite{libenLinkPredictionProblem} and Adamic-Adar \cite{adamic2003friends} are interpretable simple methods used extensively for link prediction.
More complex similarity-based methods include Katz index \cite{Katz1953}, PageRank \cite{Pageetal98}, and SimRank \cite{Jeh2002simrank}, among others.
Al Hasan \emph{et al.} \cite{Hasan06linkprediction} have explored the use of standard machine learning classifiers for link prediction.
Factorization method are also used by Menon and Elkan \cite{MenonLPfactorization} for link prediction.

Weisfeiler-Lehman Neural Machine (WLNM) \cite{zhang2017wlnm},
and Subgraphs, Embeddings, and Attributes for Link prediction (SEAL) \cite{zhang2018seal} present dedicated deep learning systems for link prediction and define the current state of the art in the space.

In weighted signed networks, for predicting edge weights, Kumar et al \cite{KumarWSN} defined the learning objective and adapted methods like Bias-Deserve \cite{Mishra2011FindingTB}, Signed Eigenvector Centrality \cite{Bonacich2007SomeUP}, PageRank \cite{Pageetal98}, and more trust based algorithms.
Edge based methods are also used in applications such as SHINE \cite{Wang_Shine} and Rev2 \cite{KumarRev2} where properties from graphs are used in determine dataset specific tasks.

For the tasks of link prediction and edge weight prediction in weighted signed networks, we present comparison of LEAP with many of these methods later in the paper.
\section{Motivation}
\label{sec-motivation}

Edges in real world networks are representative of latent properties within the graph as well as properties among the nodes.
Intuitively, for any two nodes $(u,v)$ in the graph, the properties of an edge $e_{u,v}$ between them should depend on the characteristics of the nodes themselves.
However, the nodes themselves can be characterized by the edges involving these nodes, hence increasing the dependence to neighborhoods.
Therefore, $e_{u,v}$ can be affected by several other nodes and edges in the graph and not just the node pair $(u,v)$.
In order to learn more about the edge, it is therefore important to explore the neighborhood of $u$ and $v$. 

For example, if the graph represents a professional or social network, the formation of an edge $e_{u,v}$ between $u$ and $v$ may depend on the properties of $u$ and $v$, and also on the properties of common friends and friend-of-friends, that is, on the properties of paths emanating from $u$ and $v$. 
Equally, in a trust network, the amount of trust of a user $u$ on a user $v$, constituting a label for $e_{u,v}$, can depend not only on the properties of $u$ and $v$ themselves, but also on the other sets of users that trust, and are trusted by, $u$ and $v$. 

In order to learn more about an edge $e_{u,v}$, it is therefore important to explore the neighborhood of $u$ and $v$, and assemble the nodes and edges from the graph that can impact the edge $e_{u,v}$ the most.
In particular, our framework will consider the paths originating at $u$ and ending at $v$. 
These paths can involve a large set of nodes and edges, each of which are related to the two nodes by being an intermediary in a path between them. 
These intermediate nodes and edges give us information on $u$ and $v$. 
The framework we present will enable us to learn from this shared neighborhood of $u$ and $v$ when predicting the properties of $e_{u,v}$. 

We note that we could also consider paths that originate at $u$ or $v$, but do not connect $u$ with $v$; these paths would provide a characterization of one of $u$ and $v$ only. 
It would be easy to extend our framework to consider these paths also. 
However, graph embeddings already enable us to summarize properties of individual node neighborhoods; for this reason, our framework considers mainly the {\em shared\/} neighborhood around the edge to be predicted.

\section{The LEAP Framework}
\label{sec-system}

\begin{figure}
	\centering
	\includegraphics[width=0.5\columnwidth]{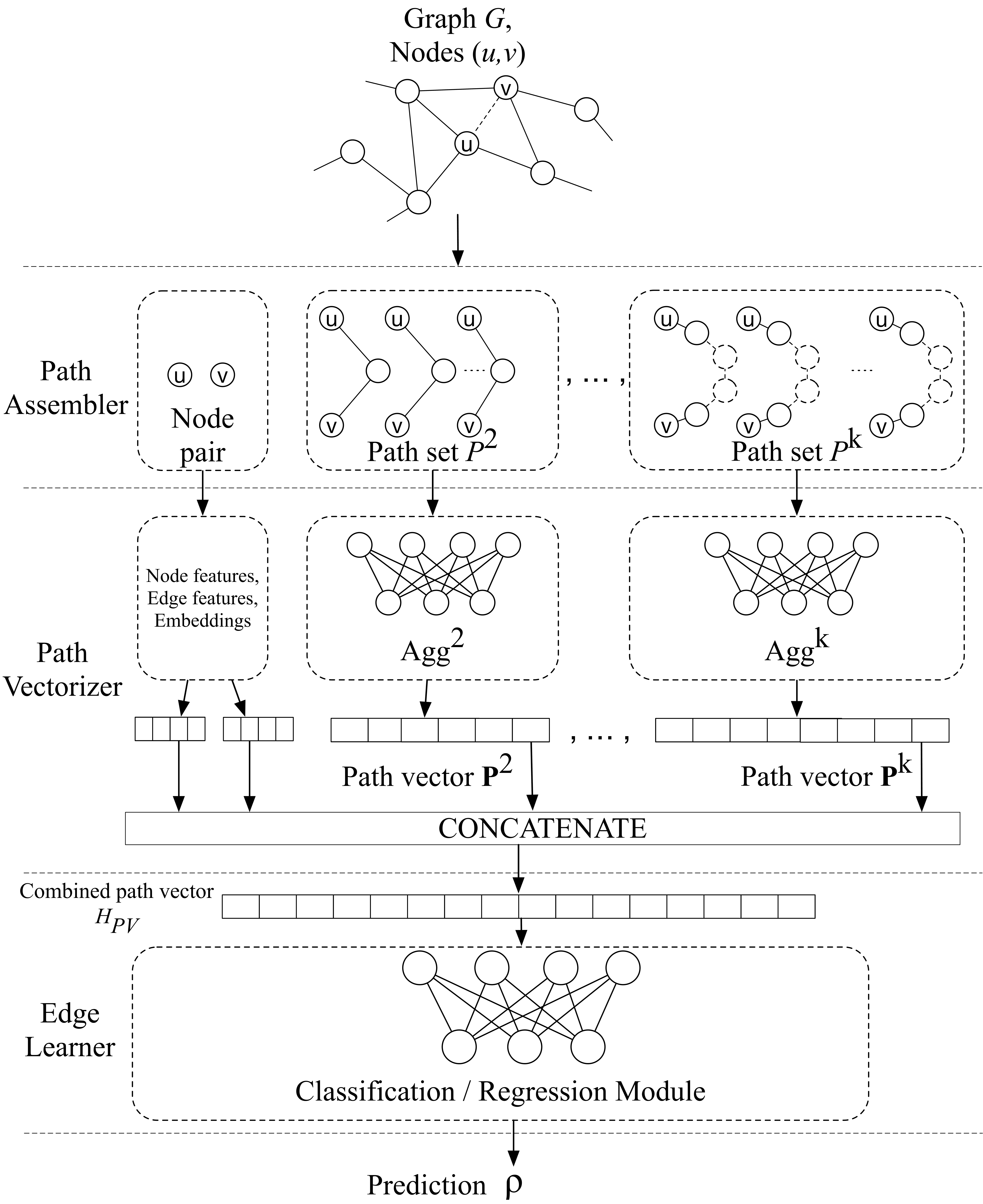}
	\caption{Overview of the LEAP framework}
	\label{fig-system}
\end{figure}

LEAP is an end-to-end deep learning framework for graph edge learning.
The core concept driving LEAP is the ability to learn edge properties in a graph simply from the graph structure, without any need for feature engineering.
Moreover, in the presence of explicit features, LEAP can use both the available features as well as self-learned representations from the structure of the graph, such as embeddings.
The LEAP framework consists of three separate modules:
path assembler, path vectorizer, and edge learner; an overview of the system is presented in Figure~\ref{fig-system}.

LEAP operates on a given graph $G=(V,E)$,
where $V$ is the set of vertices (nodes), 
and $E \subseteq V\times V$ is the set of edges.
An edge can be directed or undirected, weighted or unweighted, and signed or unsigned.
A path $p^l$ of length $l$ between the two nodes $(u,v)$ consists of a sequence of nodes $u_0, u_1, \ldots, u_l$, with $u_0 = u$ and $u_l = v$. 
The set of paths $\pathsetfull = \{p^l_1, p^l_2,\ldots,p^l_n\}$ consists of all the paths of length $l$ between the nodes $(u,v)$.
For simplicity, we will often use the notation $\path$ and $\pathset$ when referring to a path $\pathfull$ and the set of paths $\pathsetfull$ of length $l$ between the vertices $(u,v)$.

The end-to-end learning objective of LEAP is guided by the input of graph $G$, two specified nodes $(u,v)$, and the target property $\rho$  being predicted about $(u,v)$. 
For instance, in case of link prediction, where the framework can be used to predict the presence or absence of an edge between the two nodes $(u,v)$, the output prediction $\rho \in [0,1]$ from the model represents the probability of an edge existing between $(u,v)$.
Similarly, in an edge weight prediction task, the model output $\rho$ can be the predicted weight for edge $e_{u,v}$.
We now define the separate modules of the framework used for these learning objectives.

\subsection{Path Assembler}
\label{subsec-path_assembler}

The first phase of LEAP is an exploration task which gathers data from the graph structure to be used by the subsequent learning modules.
Given a graph $G$ and a pair of nodes $(u,v)$, we start by assembling paths of different lengths between the two nodes.
As the learning objective is concerned with the pair $(u,v)$, we do not include the possible 1-length path $u, v$ among the paths considered. 
As a hyper-parameter provided to the system, we define a set $\setL = \{l_1, l_2, \ldots, l_k\}$, as the set of path lengths to be used.

For any length $l \in \setL$, we now need to collect a set of paths $\pathset$ of length $l$.
For ease of computation, we can limit the size of each subset, as well as the size of $\setL$ as required by the problem and the dataset: if there are too many paths between two nodes, a random subset of paths can be extracted for use by the framework.
From the perspective of the framework, the processing is independent of these sizes.
Once collected, the paths are then made available to the system in the form of $k$ path-sets $\pathset$ with each set consisting of paths of same length $l$.

The objective of assembling the paths in length-specific sets is to allow our system to learn properties particular to path lengths.
For example, when considering paths of length $l=2$, each path between $(u,v)$ differs only by one vertex.
By processing such paths together, we increase the capability of the learning modules to focus on a specific variable property at a time, and therefore capture potential factors affecting the predicted output $\rho$.

The LEAP framework can be generalized to include the exploration of general paths from $u$, or to $v$, rather than only paths from $u$ to $v$.This can be particularly useful when few (or no) paths connect $u$ to $v$, while the nodes $u$ and $v$, individually, belong to many paths.  

\subsection{Path Vectorizer}
\label{subsec-path_vectorizer}
At the second phase of LEAP, the assembled paths are passed through a deep learning system.
This phase is called the path vectorizer.
The objective of this phase is to combine all the available information about the node pair $(u,v)$ using the nodes and the paths, and obtain a vector representation that can be used with different edge learning objectives.
This module is inspired by the deep learning architectures capable of learning from complex data structures with different dimensionality.
For a graph $G$ with $N$ nodes, each node $x \in V$ is given an integer symbol $x_i \in [1,N]$.
In order for LEAP to learn from the graph, these symbolic nodes need to be represented as vectors that can be further processed using neural networks.
Different methods of vectorization can be used to represent each node.
The simplest notation, which retains the symbolic nature of the nodes, can be to use a one-hot vector of size $N$.
In this notation, node $x_i \in V$ is represented as a vector $\chi_i$ of size $N$ where  $\chi_i[k] = 1$ if $k=x_i$, and $\chi_i[k] = 0$ otherwise.
While this representation allows one to identify each node differently, it does not provide the model with any additional information about the node that can be indicative of its relationship to other nodes or the graph structure.
For this purpose, instead of the one-hot vectors, in this paper we use the concept of dense embeddings~\cite{Mikolov2013,Perozzi2014DeepWalk,grover2016node2vec,hamilton2017inductive}.
We represent each node $x_i$ as a dense vector $\chi_i \in \mathbb{R}^K$ of a fixed dimensionality $K$.
This representation of a node, is obtained through a reference lookup $\emb$ by the entire framework, where $\chi_i = \emb(x_i)$.
Similar to the general use of dense embeddings in deep learning systems, the embeddings can either be trained with the entire system, or can be pre-trained using an embedding generation method first, and then later used with the framework.
Additionally, if we have set of node features $\calN$ representing engineered feature sets for each node, then they can be used in the framework by combining them with the embeddings.
For a node $x_i$, therefore, the embedding representation used by the system will be $\chi_i \Leftarrow \chi_i | \calN(x_i)$, where $(|)$ is a concatenation operation.

The nodes and the paths made available by the path assembler are then used into the path vectorizer using the embedding lookup $\emb$. 
Each path, represented as a sequence of nodes in their integer representation is passed through the embedding layer, in order to obtain sequences of vectors, with each sequence of length $l$ representing a path of length $l-1$.
At this stage, we have the two concerned nodes $u$ and $v$, set of path lengths $\setL$, the set of paths $\pathset$ for each length $l$, and the embedding lookup $\emb$.
In case where edge features $\calE$ are available for the graph, LEAP can use these features within the learning module as well.
We use Algorithm~\ref{alg-path_vectorizer} for obtaining a vector representation $H_{PV}$ from these inputs.

\paragraph{Aggregators.}
In Algorithm~\ref{alg-path_vectorizer}, the paths are first converted into their vector representation and then passed through the \textsc{Aggregator}.
Aggregators are the primary learning units in LEAP, and are discussed in detail in Section~\ref{sec-model}.
For each path length $l$, an associated Aggregator $Agg_l$ is responsible for processing the paths and learning a vector representation from them.
Initially, in the path vectorizer, path sets of different lengths are provided.
Since we aim at learning length-wise significant information from these paths, we process the set for each length separately.
An aggregator $Agg_l$ is a deep learning module which takes the path set $\pathsetbf$ as an input, and learns a vector representation $\hlvec$.
Due to this separate modular structure, LEAP can use several different types of aggregators.
The aggregator can also use an edge feature set $\calE$ when engineered features for the graph edges are also available.
In the path vectorizer, the vectors $\hlvec$ learned for each length $l \in \setL$ are concatenated together along with the vector representations for the input node pair $(u,v)$ to obtain a final vector representation $H_{PV}$.

\begin{algorithm}
	
	\caption{\textsc{PathVectorizer}}
	\begin{algorithmic}
		\STATE {\bfseries Input:} Node pair $(u,v)$, Path lengths $\setL$, \\
		Path sets $\pathset$ for $l \in \setL$, Embedding lookup $\emb$, \\
		Node features $\calN$, Edge features $\calE$ \\
		
		\STATE $\mathbf{u} \leftarrow \emb(u) | \calN(u)$
		\STATE $\mathbf{v} \leftarrow \emb(v) | \calN(v)$
		
		\FOR{$l$ in $\setL$}
			\STATE $\pathsetbf \leftarrow \emb(\pathset) | \calN(\pathset)$
			\STATE $\hlvec$ = \textsc{Aggregator}($\pathsetbf, \calE$)
		\ENDFOR
		\STATE $H_{PV}$ = \textsc{Concat}$[(\mathbf{u}, \mathbf{v}, \hlvec) | l \in \setL]$
		\RETURN $H_{PV}$
	\end{algorithmic}
	\label{alg-path_vectorizer}
\end{algorithm}

\subsection{Edge Learner}
\label{subsec-link_predictor}
The last step of LEAP is to perform problem specific learning on the edge $e_{u,v}$ between the two given nodes $(u,v)$.
The input to the edge learning module is the combined vector $H_{PV}$.
This vector can now be used by any classification or regression method for respective supervised learning problem.
For instance, in link prediction problem, where the objective is to detect the presence of an edge between the two nodes $(u,v)$, this module can be used as a binary classifier to classify between \emph{"link exists"} and \emph{"link does not exist"} by predicting a probability $\rho \in [0,1]$ of a link between $(u,v)$.
In general, for any edge based classification, whether binary or multi-class, the module can be used as a classifier with an input vector $H_{PV}$.
The edge learning module can also be used for regression problems.
For instance, in edge weight prediction, we can use the output $\rho$ from the module as the predicted weight for the edge. 
In case of signed edge weight, the same regression module can be used by allowing it to produce both positive and negative values.
In multi-class classification, the output $\rho$ can be treated as a vector $\rho \in \real^M$ for $M$ number of classes.

Since the edge learner by itself is simply a classification or a regression module, any corresponding learning algorithm can be used here.
For maintaining LEAP as an end-to-end trainable deep learning system, we use feed-forward neural networks for the edge learner.
These networks can vary in depth by increasing the number of layers, with the first layer receiving input vector $H_{PV}$ and the final layer predicting the output $\rho$.
Given the vector $H_{PV}$, and parameter $N_{EL}$ as the number of layers, the process of the edge learner is presented in Algorithm~\ref{alg-link_predictor}

\begin{algorithm}
	
	\caption{\textsc{EdgeLearner}}
	\begin{algorithmic}
		\STATE {\bfseries Input:} Combined Path vector $H_{PV}$, Layer count $N_{EL}$,\\
		
		\STATE $\mathbf{h_{EL}} \leftarrow H_{PV}$
		
		\FOR{$c$ = $1 \ldots N_{EL}$}
			\STATE $\mathbf{h_{EL}} \leftarrow \sigma_c(\wght_c \cdot \mathbf{h_{EL}} + \bias_c)$
		\ENDFOR
		\STATE $\rho = \sigma_p(\wght_p \cdot \mathbf{h_{EL}} + \bias_p)$
		\RETURN $\rho$
	\end{algorithmic}
	\label{alg-link_predictor}
\end{algorithm}
 
\noindent
For each layer $c$ = $1 \ldots N_{EL}$, the weight matrix of the neural network layer is represented by $\wght_c$ and the bias of the layer is represented by $\bias_c$.
$\sigma_c$ refers to the activation function used by the network layers such as $tanh$, $ReLU$, $sigmoid$ among others.
For the final prediction output layer, the weight $\wght_p$ and bias $\bias_p$ are used.
The activation function $\sigma_f$ for this layer is decided based on the nature of the problem.
In case of binary classification, often the $sigmoid \in [0,1]$ function is used.
For multi-class classification, it is common to use the $softmax$ function.
In case of normalized edge weight prediction in signed networks, with $w_{u,v} \in [-1,1]$, we use the $tanh \in [-1,1]$ activation function to obtain a floating point value representing the predicted signed edge weight.
\section{Aggregation Models}
~\label{sec-model}
In the previous section, we explained the architecture of LEAP and discussed the requirement of aggregators for performing path vectorization.
The concept of using aggregators in graphs is inspired by GraphSAGE~\cite{hamilton2017inductive} where they perform node classification by learning representations for the nodes in a graph.
In our system, we have adapted the concept of aggregators for combining paths between two nodes $(u,v)$ and obtaining representations for edges in the graph.
Within the path vectorization module, an aggregator $Agg_l$ for paths of length $l$ gets the vectorized path set $\pathsetbf$ as input with an objective of generating output vector $\hlvec$.

Since the LEAP framework consists of neural network based layers, each aggregator is itself a deep learning model where the input is a tensor of rank 4 --- (batch size, number of paths, path length, node embedding).
Each path itself is a sequence of node vectors, and each path set is a set of several such sequences.
Therefore, in order to derive single vector representation of the path set,
we first need to aggregate all the nodes in the path using aggregator $Agg_{node}$
and then aggregate these paths using the aggregator $Agg_{path}$.
The training of these aggregators is performed with the overall LEAP system using gradient descent based methods.
While several other variants are possible, we present four different kinds of aggregators used in this paper.

\subsection{AvgPool Aggregator}
Our first aggregator follows a simple architecture of combining different vectors together.
We call the model for this aggregator AvgPool.
This model relies only on the embeddings $\chi_i$ for each node $x_i$ under consideration.
The model does not have training parameters.

In AvgPool, $Agg_{node}$ concatenates all the node vectors along the path into a single vector.
Then, on the set of these derived path vectors, $Agg_{path}$ performs a one-dimensional average pooling operation.
The resulting vector $\hlvec \in \real^{(l+1)K}$ is therefore a single vector obtained by averaging the paths between the two nodes $(u,v)$ across the paths $k \in K$.
The AvgPool aggregator can be summarized as:
\begin{equation}
	\hlvec = \AvgPool([\flatten{(\pathvec_i)}, ~\forall \pathvec_i \in \pathsetbf])
\end{equation}
where $\AvgPool$ is the one-dimensional average pooling operation, and
$\flatten{( \cdot )}$ is the vector concatenation operation which combines multiple vectors
by concatenating them together.
\begin{equation}
	\flatten{([\chi_1, \chi_2, \ldots, \chi_l])} = \chi_1 | \chi_2 | \ldots | \chi_{l+1}
\end{equation}
where $(|)$ is the concatenation of two vectors.
$\pathsetbf$ is the set of vectorized paths of length $l$ in graph $G$ between the two nodes $(u,v)$.
$\pathvec_i$, indexed by $i$ is an individual path of length $l$ in set $\pathsetbf$.

AvgPool relies on the embeddings for each node, and represents a path as a fixed size vector of all the nodes combined.
Since the first and last node of these paths are the nodes $u$ and $v$ respectively, the only changing bits belong to the nodes within the paths.
By performing a bitwise pooling operation over these nodes, we can derive mean vector representations for the changing nodes in the path set.
Since the embeddings themselves are still trained by the complete framework, the gradients obtained for updating the node embeddings correspond to these average representations and their influence on the final output $\rho$.

\subsection{DenseMax Aggregator}

The DenseMax aggregator is a learning model that uses a dense (feed-forward) neural network layer for each path.
Similar to AvgPool, in DenseMax, $Agg_{node}$ obtains the representation for each path by concatenating the node vectors into a single long vector.
In this model, at $Agg_{path}$, the path vector is first passed through a dense neural layer.
The resulting activations are then passed through a max-pooling operation which helps derive a single vector representation for the paths of length $l$.
Therefore, $Agg_{path}$ in DenseMax consists of a dense neural network layer and a one-dimensional max pooling operation.
The operations of the DenseMax aggregator can be summarized as:
\begin{equation}
	\hlvec = \MaxPool([\sigma(\wght_l \cdot \flatten{(\pathvec_i)} + \bias_l), \forall \pathvec_i \in \pathsetbf])
\end{equation}
where $\wght_l$ and $\bias_l$ are the weight matrix and bias for the dense neural layer.
$\sigma$ is the activation function used by the dense layer.
$\flatten{( \cdot )}$ is the vector concatenation operation.
$\MaxPool$ is the one-dimensional max pooling operation which selects bitwise maximum value from multiple vectors to derive a single final vector.
The $\MaxPool$ operation is used on these representations in order to capture the most activated bits that can affect the final output $\rho$.

\subsection{SeqOfSeq Aggregator}

The sequence of nodes from $u$ to $v$ can hold information relevant to the final prediction.
For instance, if the existence of an edge between $u$ and $v$ depends on the presence of a path between the two nodes with consecutively increasing edge weight, the sequential order of nodes contains information which holds significance to the final outcome of the model.
Therefore, in SeqOfSeq aggregator, we treat the paths as ordered sequences of nodes.
We can further consider the path set as a sequence of different paths, if the paths can be ordered using some characteristics.
For example, if the edges are labeled by weights, then the total weight of a path $\path$ is the sum of edge weights for each edge in the path.
If the paths from $u$ to $v$ are then sorted according to their total weights, we can process them in a specified order.
To this end, the aggregators $Agg_{node}$ and $Agg_{path}$ would need to be sensitive to the order of the inputs.

In the SeqOfSeq aggregator, we first use an LSTM $\lstm_{inner}$ on each path. 
From the output activations of $\lstm_{inner}$, we extract a vector representation for the path by performing a max-pooling operation.
The aggregator $Agg_{node}$, in this case, consists of both the $\lstm_{inner}$ and the max-pooling operation.
We use a max pool here instead of using only the activation from last timestep of the $lstm_{inner}$:
we believe that since our objective is to extract information from the path itself, a max-pooling operation can be more effective in summarizing the path than the final activation.
After summarizing each path into a single vector, the sequence of path vectors is processed by a combination of another LSTM $\lstm_{outer}$, followed by a max-pooling operations, as the $Agg_{path}$ aggregator.
The SeqOfSeq aggregator can be summarized as:
\begin{equation}
	\begin{split}
	& H_{inner} = [\MaxPool(\lstm_{inner}(\pathvec_i)), \forall \pathvec_i \in \pathsetbf]\\
	& \hlvec = \MaxPool(\lstm_{outer}(H_{inner}))
	\end{split}
\end{equation}
where $\MaxPool$ is the one-dimensional max pooling operation,
$H_{inner}$ is the intermediate sequence of derived path vectors,
and $\lstm_{inner}$ and $\lstm_{outer}$ are the inner and outer LSTMs respectively,
used for processing corresponding sequences.

Variants of SeqOfSeq can also be created to use order information only at the paths, or only at the nodes.
With the use of sequence learning neural networks such as LSTMs, the SeqOfSeq aggregator is more powerful than the AvgPool or DenseMax aggregators, and it trains a much larger number of parameters.

\subsection{EdgeConv Aggregator}

Edges of a path can themselves contain significant information.
In order to emphasize learning also from the the edges, we propose an aggregator called EdgeConv which focuses on edges while operating over paths.
In order to build a learning widget that can operate on the edge, we use a one-dimensional Convolutional Neural Network (CNN) with a window size of 2 that takes as input two consecutive nodes forming an edge.
Therefore, when a path is represented as a sequence of nodes, the convolution kernel focuses on all pairs of consecutive nodes along the edge.
Given the convolutional results on all pairs of consecutive nodes, we apply a max-pooling operation to compute the overall path label.
The aggregator $Agg_{node}$ for EdgeConv, therefore, consists of a one-dimensional CNN and a max-pooling operation.
Considering the set of derived paths as an ordered sequence, the $Agg_{path}$ for this case also uses an LSTM and max-pooling operation.
Therefore, all the path vectors for paths of length $l$ derived using $Agg_{node}$ are then processed using an LSTM, followed by another max-pooling operation to derive the final vector representation $\hlvec$.
EdgeConv can be summarized as :
\begin{equation}
	\begin{split}
	& H_{inner} = [\MaxPool(\convOne(\pathvec_i)), \forall \pathvec_i \in \pathsetbf]\\
	& \hlvec = \MaxPool(\lstm(H_{inner}))
	\end{split}
\end{equation}
where $\MaxPool$ is the one-dimensional max pooling operation,
$H_{inner}$ is the intermediate sequence of derived path vectors,
and $\lstm$ is the LSTM module used to learn from different path vectors.

Similar to SeqOfSeq, it is not necessary to treat the paths as an ordered sequence, and different variants of EdgeConv can consider the set of derived paths as an unordered set.
The sequence of nodes, however, needs to be ordered in EdgeConv, as it operates consecutively over the edges.

\subsection{Aggregator Extensions}

The four aggregators presented above provide different levels of complexity and use different neural network modules.
Together, they illustrate how the modular nature of LEAP allows us to use different neural network architectures as part of the full system for an end-to-end training.
We believe that the use of aggregators is key to achieving an extensible and flexible framework.
Specialized aggregators can be trained by focusing on significant properties under concern.
The aggregators can also include the new deep learning concepts of attention~\cite{bahdanauAttention,luongAttention} and memory~\cite{Weston2014}.
Similarly, graph specific neural models like Graph Convolutional Networks~\cite{kipf2017semi} and Graph Attention Networks~\cite{gat2018} can be adapted as aggregators by representing the assembled path sets as subgraphs.
While we train the aggregators along with the entire framework, they can be trained separately using any objective function.
In case of transfer learning, a well trained model can be transfered into the LEAP framework and can be used simply as a function without training further.
Similarly, a partially trained model can be used as an aggregator, and it can further be trained by the learning objective of the complete framework.
\section{Evaluation}
\label{sec-eval}
The design of the LEAP system, and the use of aggregators, make it an easy to use end-to-end learning system for any kind of graph.
For small graphs with fewer nodes, simpler aggregators with few parameters can be used.
For very large datasets, we can construct complex aggregators targeted at several latent properties in the graph.
In order to demonstrate the learning abilities of this system, we evaluate it on two commonly studied problems in graphs and social networks --- link prediction, and edge weight prediction in weighted signed networks.

\def\arraystretch{1.2}
\begin{table}[]
	\small
	\centering
	\caption{Summary of the datasets used for evaluation}
	\begin{tabular}{llll}
		\hline
		\textbf{Type}    & \textbf{Name} & \textbf{Nodes} & \textbf{Edges} \\\hline
		Link Prediction & USAir         & 332            & 2,126           \\
		& NS            & 1,589           & 2,742           \\
		& PB            & 1,222           & 16,714          \\
		& Yeast         & 2,375           & 11,693          \\
		& C.ele          & 297            & 2,148           \\
		& E.coli         & 1,805           & 14,660          \\
		& arXiv         & 18,722          & 198,110         \\
		& FB            & 4,039           & 88,234          \\\hline
		Weighted Signed & Bitcoin-OTC & 5,881 & 35,592\\
		Networks& Bitcoin-Alpha & 3,783 & 24,186\\
		& Wikipedia-RFA & 9,654 & 104,554\\\hline
	\end{tabular}
	\label{tab-datasets}
\end{table}

\subsection{Link Prediction}
Graphs and networks evolve over time by creation of newer links between the nodes.
Given a graph $G$ and a pair of nodes $(u,v)$, the link prediction problem aims at predicting the probability of existence of an edge $e_{u,v}$ between the two nodes.

\subsubsection{Learning Objective}
In order to learn this objective using LEAP framework, in the Edge Learner module, we can treat it as a binary classification problem.
From the graph $G = (V,E)$, the set of edges $E$ can be considered as the positive sample set.
Similarly a set of node pairs $(x_1,x_2) \in V$ can be sampled from the graph where edge $e_{x_1,x_2} \notin E$ is the negative set for classification. 
A label $\tau=1$, therefore can be associated with positive pairs and a label $\tau=0$ is associated with the negative pairs.

\subsubsection{Datasets}
We evaluate the LEAP Link Prediction model on eight real world datasets.
The choice of datasets is motivated by the link prediction results presented in two state of the art models --- WLNM~\cite{zhang2017wlnm} (Weisfeiler-Lehman Neural Machine) and SEAL~\cite{zhang2018seal} (Subgraphs, Embeddings, and Attributes for Link prediction).
The datasets used in this paper are listed in Table~\ref{tab-datasets}.

USAir~\cite{usair} is a network graph for US airlines.
Network Science (NS)~\cite{Newman2006FindingCS} is the collaboration network for researchers in the subject of network science.
Political Blogs (PB)~\cite{politicalBlogs} is a political blog network form the US.
Yeast~\cite{yeast} is a PPI (protein-protein interaction) network for yeast.
C.ele is the neural network of the worm Caenorhabditis elegans~\cite{Watts1998Collective}.
E.coli is the dataset of pairwise reaction network of metabolites in E.coli~\cite{Zhang2018BeyondLP}.
arXiv~\cite{snapnets} is the collaboration network of research papers on arXiv under the Astro Physics category.
Facebook (FB)~\cite{snapnets} is the dataset of friend lists from the Facebook social network.

\def\arraystretch{1.2}
\begin{table*}[]
	\centering
	\caption{Area under the ROC curve (AUC) comparison of LEAP with baselines.
		Best LEAP results and best dataset results are highlighted.
		\emph{n2v} refers to the use of node2vec embeddings with LEAP. \emph{OOM} refers to Out-of-Memory.}
	\begin{tabular}{lllllllll}
		\hline
		& USAir  & NS     & PB     & Yeast  & C.ele  & E.coli & arXiv  & FB     \\\hline
		Adamic-Adar         & 0.9507 & 0.9498 & 0.9250 & 0.8973 & 0.8659 & 0.9524 & -      & -      \\
		Katz                & 0.9273 & 0.9524 & 0.9306 & 0.9264 & 0.8606 & 0.9329 & -      & -      \\
		PageRank            & 0.9486 & 0.9529 & 0.9374 & 0.9314 & 0.9046 & 0.9548 & -      & -      \\\hline
		node2vec            & 0.9122 & 0.9198 & 0.8621 & 0.9407 & 0.8387 & 0.9075 & 0.9618 & 0.9905 \\
		Spectral Clustering & 0.7482 & 0.8829 & 0.8261 & 0.9346 & 0.5007 & 0.9514 & 0.8700 & 0.9859 \\\hline
		WLK                 & 0.9598 & 0.9864 & OOM    & 0.9550 & 0.8965 & OOM    & -      & -      \\
		WLNM                & 0.9571 & \textbf{0.9886} & 0.9363 & 0.9582 & 0.8603 & 0.9706 & 0.9919 & 0.9924 \\
		SEAL                & \textbf{0.9729} & 0.9761 & 0.9540 & \textbf{0.9693} & 0.9114 & 0.9704 & 0.9940 & 0.\textbf{9940} \\\hline
		LEAP-AvgPool         & 0.9259 & 0.9362 & 0.9555 & 0.9474 & 0.9011 & 0.9484 & 0.9918 & 0.9916 \\
		LEAP-DenseMax       & 0.9555 & \textbf{0.9785} & 0.9541 & 0.9573 & 0.9050 & 0.9662 & 0.9940 & 0.9914 \\
		LEAP-SeqOfSeq       & 0.9576 & 0.9635 & 0.9547 & 0.9540 & 0.9153 & 0.9626 & \textbf{0.9941} & 0.9907 \\
		LEAP-EdgeConv       & \textbf{0.9639} & 0.9621 & \textbf{0.9577} & 0.9554 & 0.9058 & 0.9614 & \textbf{0.9941} & 0.9908 \\\hline
		LEAP-n2v-AvgPool     & 0.9086 & 0.9068 & \textbf{0.9586} & 0.9551 & 0.8909 & 0.9505 & 0.9919 & 0.9920 \\
		LEAP-n2v-DenseMax   & 0.9518 & 0.9636 & 0.9564 & 0.9652 & 0.9129 & \textbf{0.9719} & 0.9934 & 0.9914 \\
		LEAP-n2v-SeqOfSeq   & 0.9532 & 0.9618 & 0.9571 & 0.9610 & 0.9083 & 0.9662 & 0.9938 & \textbf{0.9924} \\
		LEAP-n2v-EdgeConv   & 0.9547 & 0.9622 & 0.9575 & \textbf{0.9639} & \textbf{0.9185} & 0.9678 & \textbf{0.9941} & 0.9921
	\end{tabular}
		\label{tab-results}
\end{table*}

\subsubsection{Experiment Setup}
We performed an extensive set of experiments on the above mentioned datasets in order to evaluate and compare our framework against state of the art methods in link prediction.
For each dataset, we sampled a variable number of data samples into the training set and evaluated the model on the remaining samples.
The results presented in this paper adopt the partitioning used in the current state of the art method SEAL~\cite{zhang2018seal}.
For smaller datasets with less than 2500 nodes, we use 90\% of the graph edges and an equal number of negative samples for training, and present evaluation results on the remaining 10\% edges and equal number of negative pairs.
For relatively larger datasets with more than at least 4000 nodes, we partition the training and evaluation datasets at 50\%.

The LEAP\footnote{The code is available at \url{https://github.com/rakshit-agrawal/LEAP}} system and the aggregators were written in Python
with Keras~\cite{chollet2015keras} deep learning framework,
using the Tensorflow~\cite{tensorflow2015} backend.
The hyperparameters for each aggregator were selected using multiple trials.
We report the results with the best hyperparameters for each setting individually.
In all the reported results, we used the path lengths in set $\setL = \{3,4\}$,
and used upto 50 paths for each length selected randomly.
All the methods were trained using a loss function of binary cross-entropy and the Adam~\cite{Kingma2014AdamAM} optimizer for gradient descent with a learning rate of 0.001.
Each model was trained for upto 30 epochs with early stopping enabled.

\begin{figure*}[tbh]
	\centering
	\includegraphics[width=0.32\linewidth]{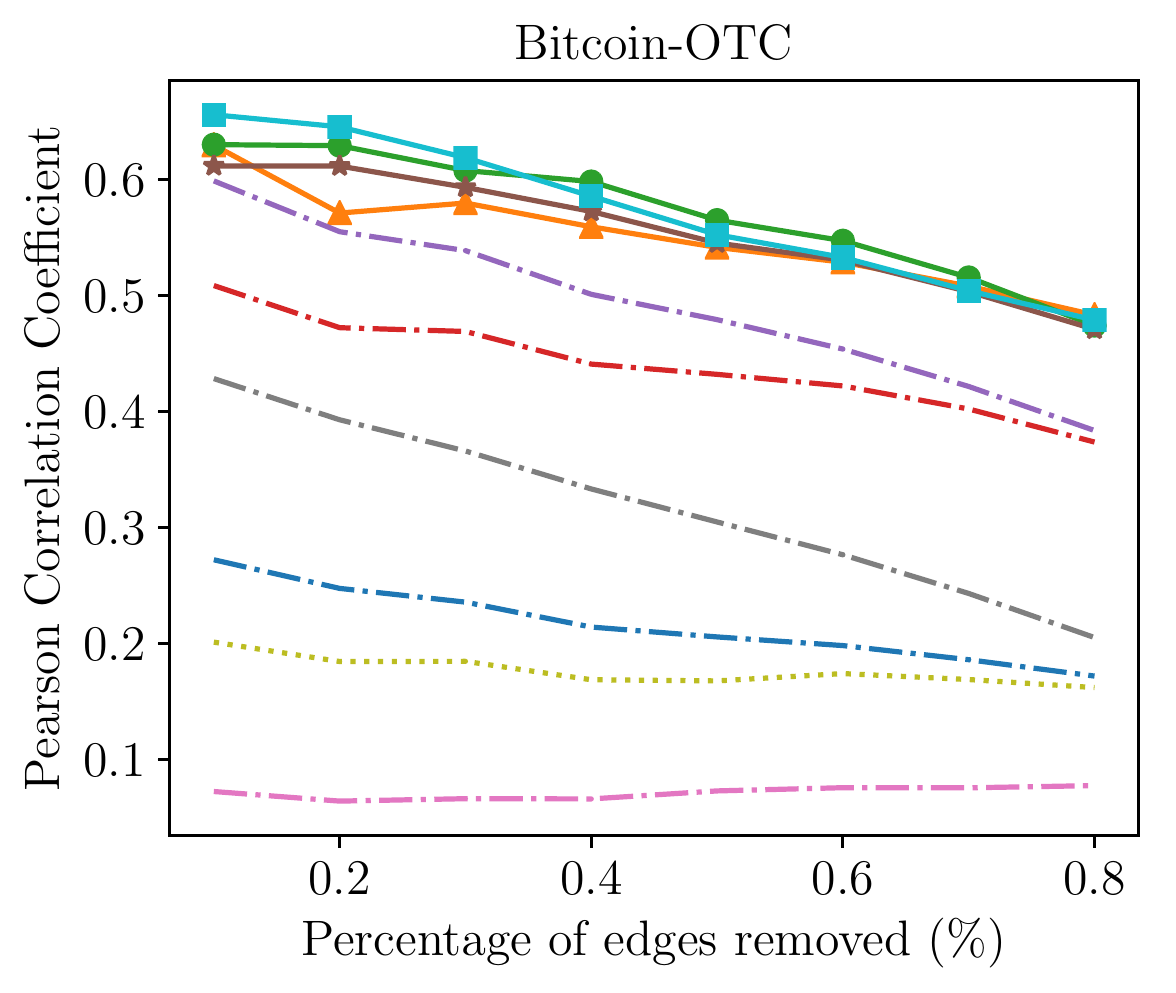}
	\includegraphics[width=0.32\linewidth]{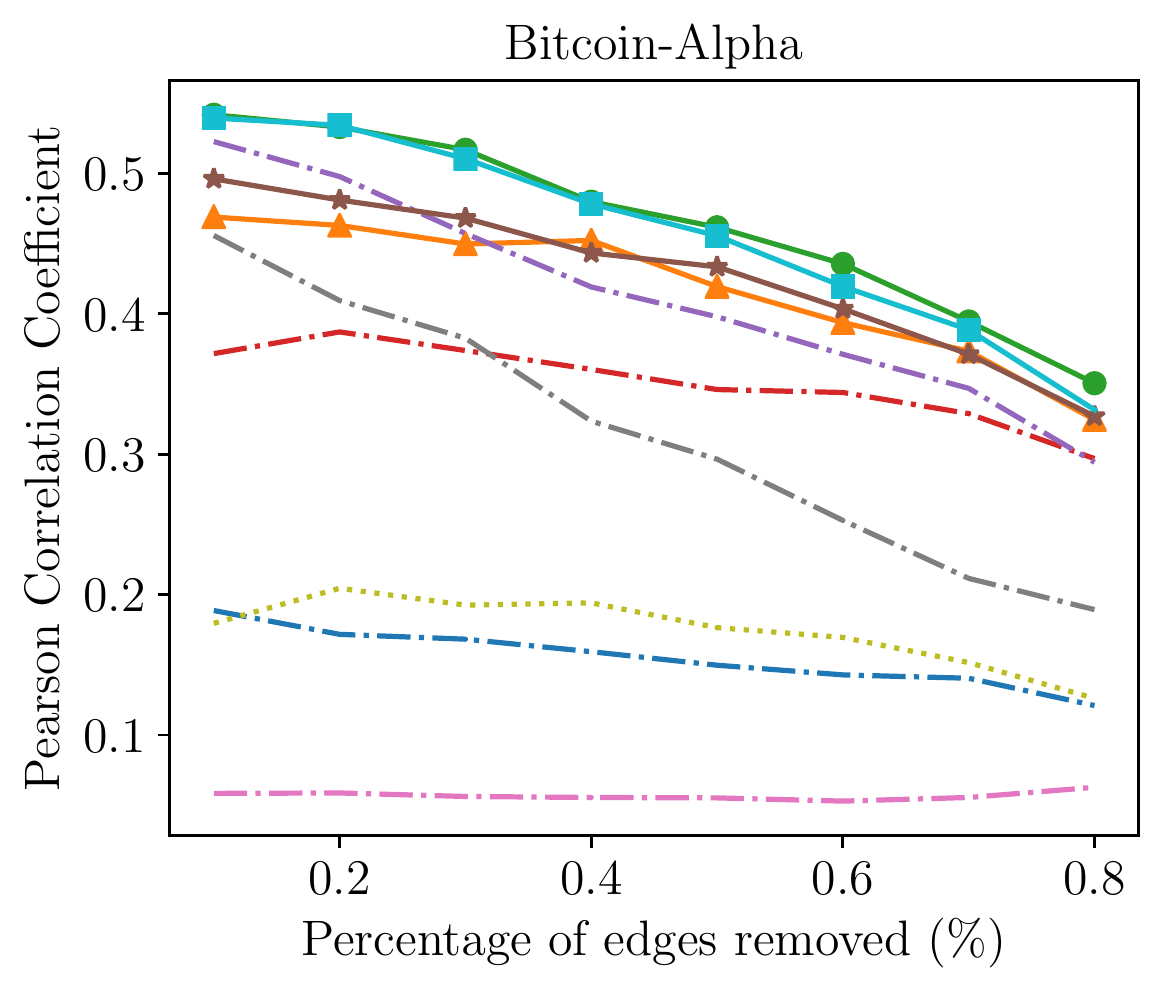}
	\includegraphics[width=0.32\linewidth]{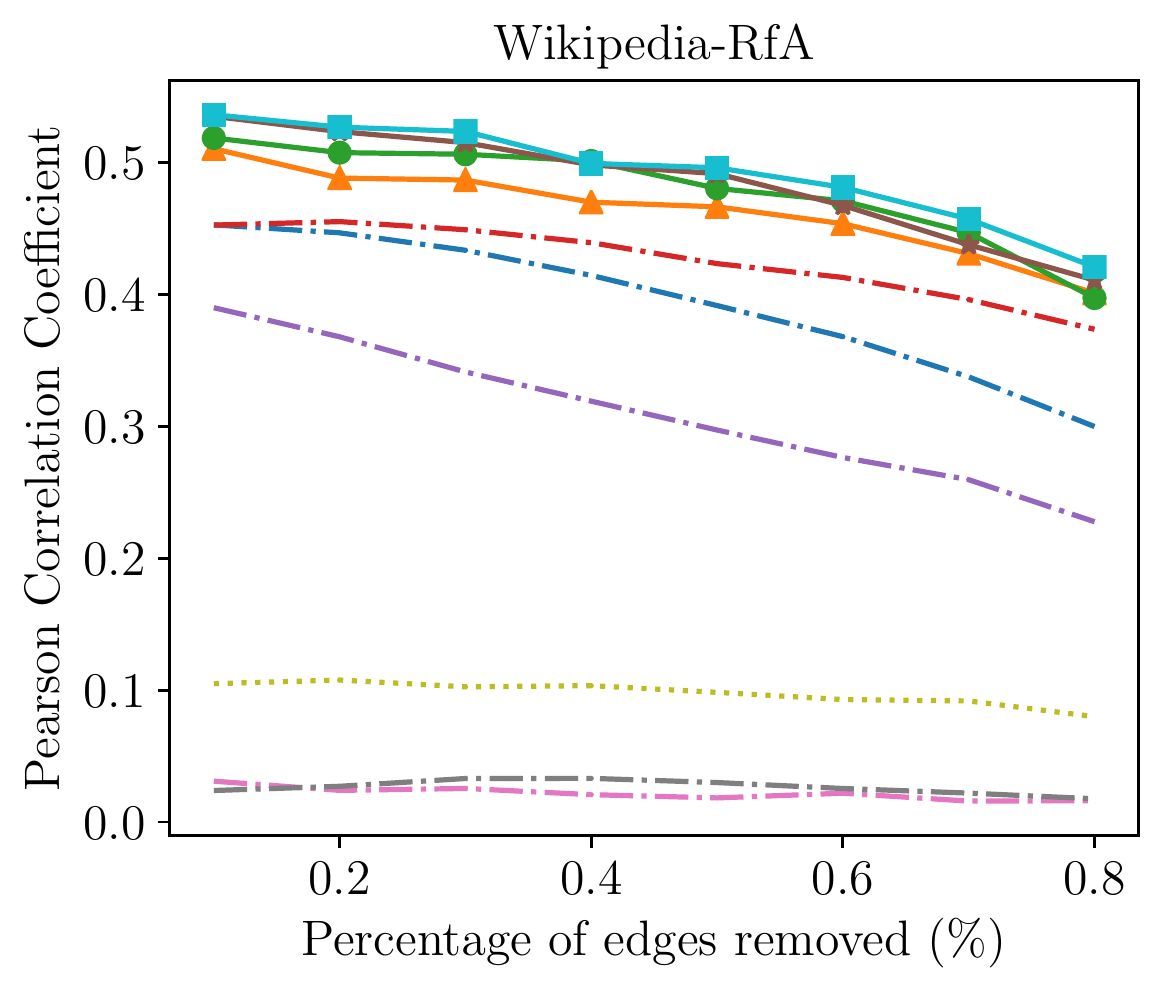}
	\includegraphics[width=0.32\linewidth]{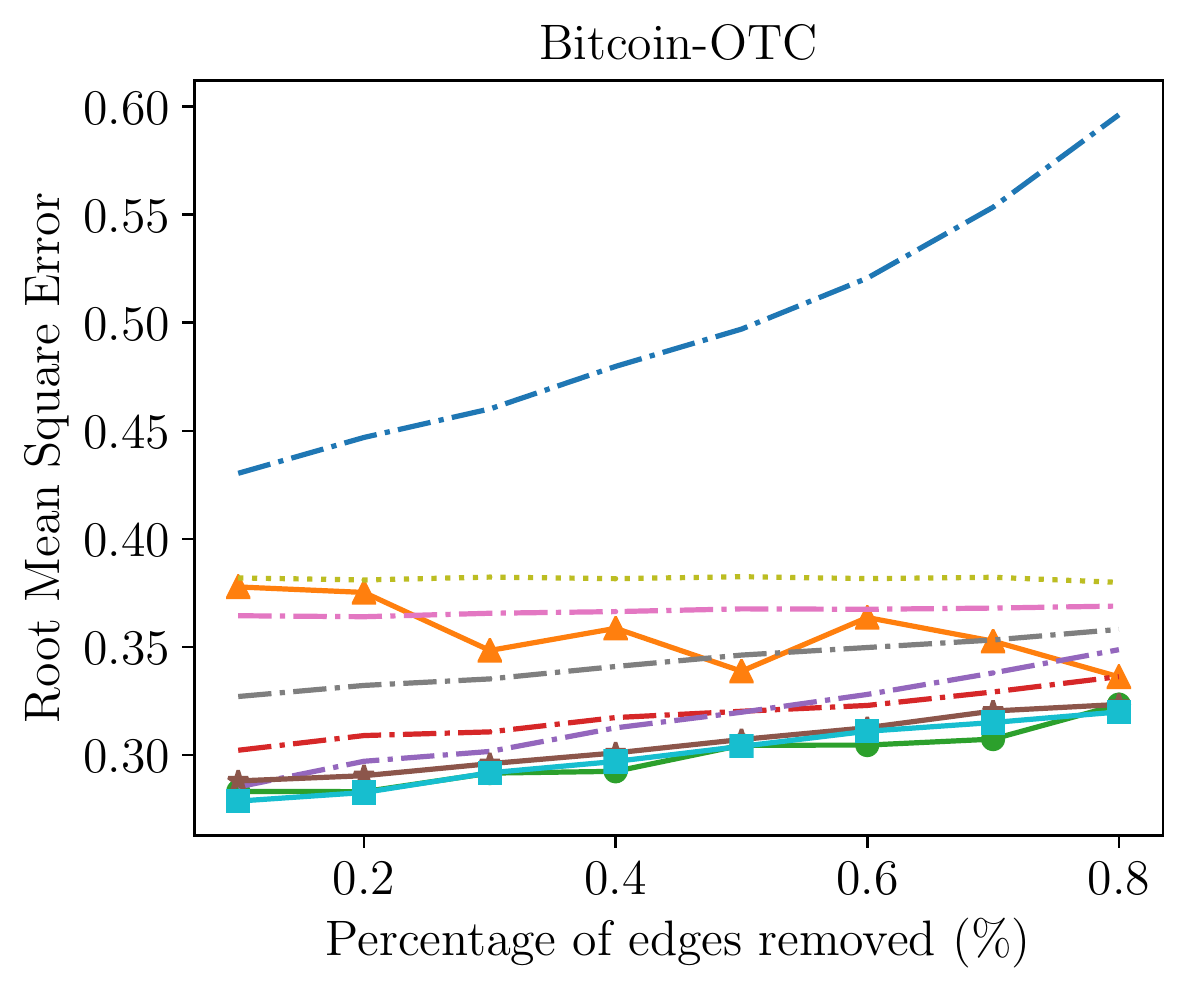}
	\includegraphics[width=0.32\linewidth]{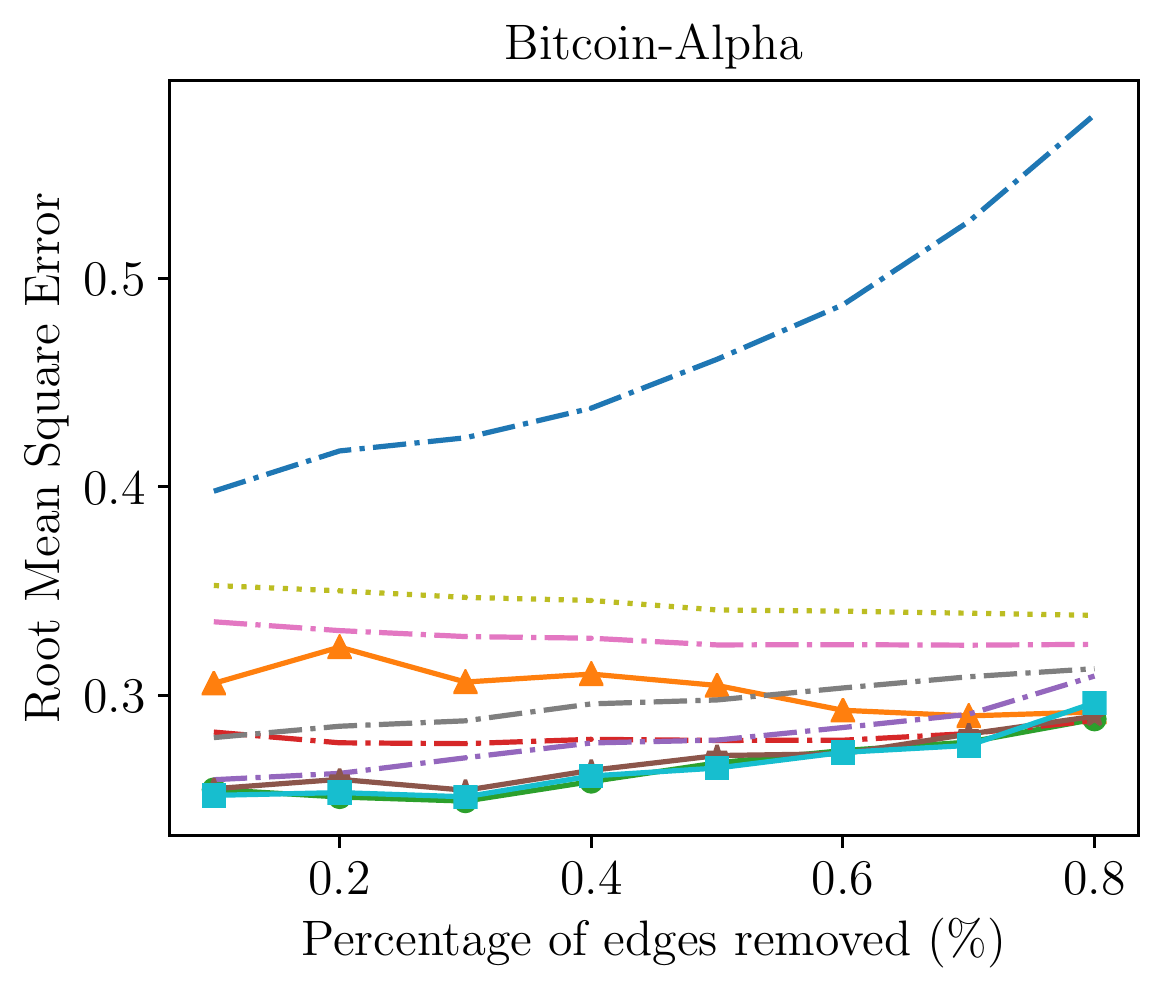}
	\includegraphics[width=0.32\linewidth]{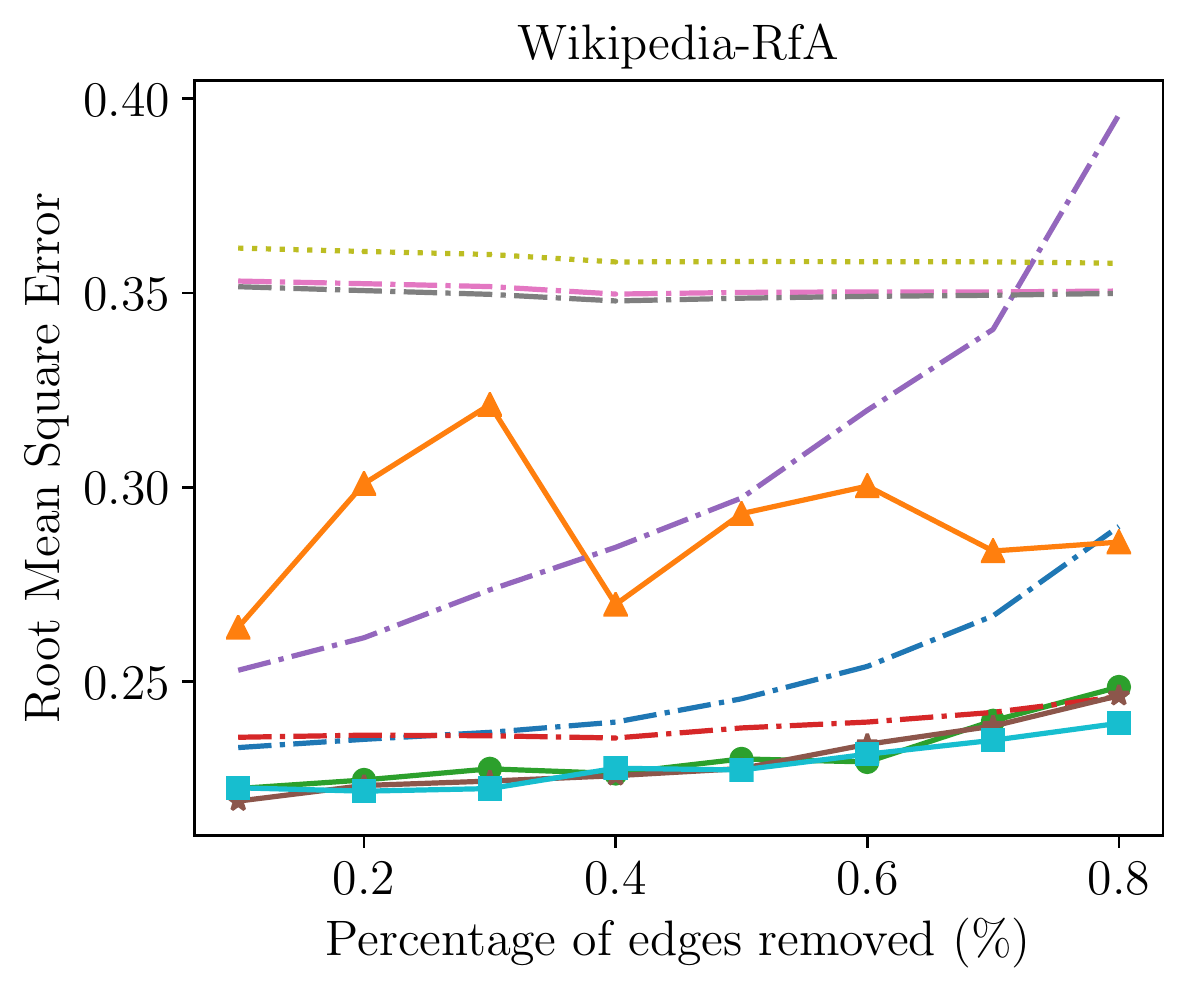}
	\includegraphics[width=0.99\linewidth]{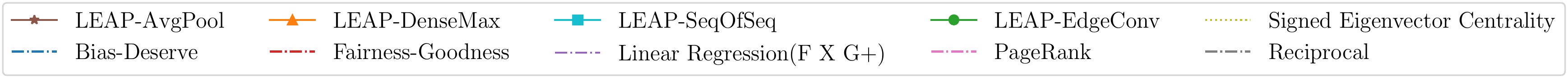}

	\caption{Plots for PCC(top) and RMSE(bottom) on the three datasets for Weighted Signed Networks. The x-axis refers to the percentage of edges removed while training the models. Over all the datasets, and along both the metrics, LEAP based methods show significantly better performance, with the complex aggregators SeqOfSeq and EdgeConv giving the best performance on both the metrics.}
	\label{fig-wsn-plots}

\end{figure*}

\subsubsection{Results}
We present the results obtained as per the above mentioned experimental setup for LEAP in table~\ref{tab-results}.
The results comprise of LEAP with the four aggregators --- AvgPool, DenseMax, SeqOfSeq, and EdgeConv, discussed in the paper.
Additionally, we also compare the ability of our model to learn node embeddings themselves against using pre-trained embeddings.
For pre-trained embeddings, similar to SEAL, we use the node2vec~\cite{grover2016node2vec} method first on the graph to derive node embeddings, and then use them with the LEAP framework without updating them further during the system training.
We compare these results with three different kinds of methods used for link prediction.
We first use the heuristics including Adamic-Adar~\cite{adamic2003friends}, Katz index~\cite{Katz1953}, and the PageRank~\cite{pagerank} algorithm.
For feature learning based models, we use spectral clustering~\cite{Tang2011}, and the node2vec~\cite{grover2016node2vec} algorithm which learned node embeddings and then performs a link prediction task on them.
Finally, we compare our system with subgraph-based link prediction methods, defining the current state of the art.
These methods include Weisfeiler-Lehman graph kernel (WLK)~\cite{nino2011wlk}, Weisfeiler-Lehman Neural Machine (WLNM)~\cite{zhang2017wlnm}, and Subgraphs, Embeddings, and Attributes for Link prediction (SEAL)~\cite{zhang2018seal}.
Performance of all the models is compared using the Area under the ROC curve (AUC) metric.
We use the settings and results by Zhang et al.~\cite{zhang2018seal}
for all our baseline methods.

As can be seen in table~\ref{tab-results}, LEAP performs best or close-to-best on each dataset.
Further, we show that an external method for learning node embeddings like node2vec can easily be used within the system.
Similarly we can also incorporate known feature vectors for the nodes whenever available.

In comparison to deep learning methods and current state of the art WLNM and SEAL, LEAP achieves equivalent or better performance with the presented aggregators.
However, due to its modular nature, the LEAP framework can be highly extended and adapted for different datasets with different latent properties.
Further, SEAL requires two prominent steps of node labeling and embedding generation before the neural network can be trained on the graph.
In making a learning framework easily deployable on multiple platforms, it is highly advantageous to have an end-to-end trainable system and LEAP provides this particular ability with sufficient modularity to tune the simplicity of the model as required.

\subsection{Weighted Signed Networks}
In real world datasets representing relations of certain kind among the nodes, the edges can possess different meaningful properties that are of significance to the underlying network.
For instance, in a user-user interaction system, each user can have a trust associated with another user. 
This trust further propagates through the network, affecting the trust between two different users in a certain way.
A generalized representation of such networks is obtained through weighted signed networks (WSN).
A WSN consists of a graph $G=(V,E)$ where an edge $e_{u,v} \in E$ between two nodes $(u,v) \in V$ has a weight $w_{u,v}$ associated to it.
The weight $w_{u,v}$ can be a signed weight, specifying a positive or negative sentiment with a magnitude $|w_{u,v}|$ specifying the intensity of the relation.
A WSN graph can either be directed or undirected.

\subsubsection{Learning Objective}
Kumar et al~\cite{KumarWSN} presented the task of predicting edge weight in WSNs where given a graph $G$ and a node pair $(u,v)$, the objective is to predict the signed edge weight $w_{u,v}$ between the two nodes.
For simplicity, the edge weights are normalized to a scale of $w_{u,v} \in [-1,1]$.
In the case of the LEAP framework, the problem of predicting edge weight can be treated as a regression problem for the Edge Learner module.
From the graph $G = (V,E)$, we can use the entire set of edges $E$ with the edge nodes $(x_1,x_2) \in V$ for edge $e_{x_1, x_2} \in E$ being the input nodes to the system.
Each such node pair can be associated with a regression label $\tau = w_{x_1,x_2}$ using the weights associated with each edge.

\subsubsection{Datasets}
For evaluation of the LEAP WSN Edge Weight Predictor, we use three real world datasets of user-user interaction networks.
These datasets used here are influenced by Kumar et al~\cite{KumarWSN} and are available from SNAP database~\cite{snapnets}.
The three datasets are listed in Table~\ref{tab-datasets}.

Bitcoin-OTC~\cite{KumarWSN} is the ``who-trusts-whom'' network of people trading Bitcoins on the platform ``Bitcoin OTC''. 
The directed signed edge weight between the users on this network refers to the rating given by a user to the other user on the network on a scale from -10 to 10.
Bitcoin-Alpha~\cite{KumarWSN} is a similar network for a different trading platform called ``Bitcoin Alpha''.
Wikipedia-RFA~\cite{West2014ExploitingSN, KumarWSN} is a voting network for Wikipedia request for adminship (RfA).
The edges on the network refer to directed votes between users.
To associate weights to the signed edges, Kumar et al~\cite{KumarWSN} used the VADER sentiment engine~\cite{Hutto2014VADERAP} and used the difference between the positive and negative sentiment scores obtained for the vote explanation text in the Wikipedia-RFA dataset.

\subsubsection{Experiment Setup}
In this evaluation, we present and compare the results for regression on predicting edge weights with $\delta\%$ edges removed.
We vary the value of $\delta$ between 10\% to 80\%, with a step size of 10\%, specifying the range for partitioning the training and the evaluation datasets. 

The system and hyperparameter settings for this task are same as the link prediction model.
The only difference in this adaptation of LEAP for a regression task is the choice of loss function.
We use the Mean Squared Error (MSE) as the loss function used to compute the gradients for training the Edge Weight Prediction model.
The hyperparameter selection was performed using multiple trials, and the best settings were used to report the results.

\subsubsection{Results}

Similarly to link prediction, we evaluated our system for this objective with the four aggregators --- AvgPool, DenseMax, SeqOfSeq, and EdgeConv, discussed in the paper.
Neural network based models are not used directly for the task of edge weight prediction in WSN and therefore we compare our results to the heuristic and feature learning based methods previously applied on these datasets.

In the definition of edge weight prediction on WSNs by Kumar et al~\cite{KumarWSN},
they adapted several algorithms for this task, providing us with a set of baseline measures on these datasets.
We first use a basic method of Reciprocal~\cite{KumarWSN}, where the edge weight $w_{u,v}$ is same as that of the reciprocal edge weight $w_{v,u}$ if there exists an edge $e+{v,u} \in E$, and 0 otherwise.
We then use two graph algorithms PageRank~\cite{Pageetal98} and Signed Eigenvector Centrality~\cite{Bonacich2007SomeUP}.
Each of these algorithms independently learn a score for each node in the graph.
The edge weight predicted by these methods simply refers to the difference between the scores obtained for the nodes $u$ and $v$.
Finally, we compare our system with relation specific algorithms used for extracting interaction measures between the nodes.
These include Bias-Deserve~\cite{Mishra2011FindingTB} and Fairness-Goodness~\cite{KumarWSN}.
These are iterative algorithms that associate two properties with each node, and learn them by performing sequential iterations on all the nodes in the graph, updating one property at a time.
An additional method used in~\cite{KumarWSN} is a Linear Regression model using the above mentioned values as features. 
This method is identified as Linear Regression (F $\times$ G +)

All these methods were evaluated using the experiment setup mentioned above and the results were measures across two standard metrics.
We first measure the Root Mean Square Error (RMSE) on the predicted weights to highlight the closeness between predicted and true weights.
We then measure the Pearson Correlation Coefficient (PCC) on the predicted weights in order to measure the relative trend in the prediction.

As can be seen in Figure~\ref{fig-wsn-plots}, LEAP based methods outperform all the baseline methods on the three datasets, hence defining the new state of the art for edge weight prediction.
Moreover, with decreasing number of known edges, the performance of LEAP degrades much slower compared to the other methods.

\subsection{Discussion}
The results presented above on two edge learning problems prove the ability of LEAP to learn edge properties directly from the structure of the graph.
Beyond this, we believe that this framework presents a unique quality of extensibility.
LEAP is a highly modular system that can be adapted for any learning objective on the egde properties.
In the two tasks presented in the paper, by just making a small change in the edge learner module for each task, the system was able to match the state of the art results.
In doing so, it did not require any heuristics or feature extraction methods, and could learn only through the aggregation of paths between the two concerned nodes.
Further, LEAP can be trained end-to-end, making it much more convenient to code, use, and maintain, as well as more interpretable since the weights throughout the network are updated using a common objective.
The adaptability of LEAP also makes it a potential platform for future research in learning edge properties, and for use of neural networks in graphs.
\section{Conclusions}
\label{sec-conc}
We presented a novel end-to-end deep learning framework called LEAP for learning edge properties in a graph in this paper.
LEAP includes a modular structure consisting of a path assembler, path vectorizer, and an edge learner.
For any graph $G=(V,E)$ and two given nodes $(u,v) \in V$, the system learns properties associated with the edge $e_{u,v}$ by aggregating paths between them from the graph.
The aggregation is performed using deep learning modules which can be selected based on the dataset and the problem under concern.
The system can perform different kinds of supervised learning tasks such as binary or multi-class classification, multi-label classification and regression among others.
Being powered by neural modules, the complete framework of LEAP is a layered deep learning model that can be trained end-to-end using gradient descent based methods.

We demonstrate that LEAP can obtain state-of-the-art performance for different learning problems on several real world datasets.
For two specific problem of link prediction, and edge weight prediction in weighted signed networks, LEAP shows great performance by matching or improving upon the current state of the art.
We also show that the LEAP framework is easily extensible, and can also incorporate node embeddings, node features and edge features into the system.
We believe that this system can act as a great platform for experimentation in edge learning, and can be adapted for several different problems.
We also believe that the simple architecture of LEAP allows it to be an easily deployable neural model in production environments.

\bibliographystyle{alpha}
\bibliography{pathlinks}

\end{document}